\documentclass[preprint,12pt]{elsarticle}

\usepackage{amssymb}
\usepackage{hyperref}
\usepackage{tikz}
\usepackage{xcolor}

\journal{arXiv}

\begin{document}

\begin{frontmatter}




\title{From ``um'' to ``yeah'': Producing, predicting, and regulating information flow in human conversation}


\author[inst1]{Claire Augusta Bergey}
\author[inst2,inst3]{Simon DeDeo}
\affiliation[inst1]{organization={Department of Psychology},
            addressline={University of Wisconsin-Madison}, 
            city={Madison},
            postcode={53706}, 
            state={WI},
            country={USA}}
\affiliation[inst2]{organization={Department of Social and Decision Sciences},
            addressline={Carnegie Mellon University}, 
            city={Pittsburgh},
            postcode={15213}, 
            state={PA},
            country={USA}}
\affiliation[inst3]{organization={Santa Fe Institute},
            addressline={1399 Hyde Park Rd}, 
            city={Santa Fe},
            postcode={87501}, 
            state={NM},
            country={USA}}

\begin{abstract}
Conversation demands attention. Speakers must call words to mind, listeners must make sense of them, and both together must negotiate this flow of information, all in fractions of a second. We used large language models to study how this works in a large-scale dataset of English-language conversation, the CANDOR corpus. We provide a new estimate of the information density of unstructured conversation, of approximately 13 bits/second, and find significant effects associated with the cognitive load of both retrieving, and presenting, that information. We also reveal a role for \emph{backchannels}---the brief yeahs, uh-huhs, and mhmms that listeners provide---in regulating the production of novelty: the lead-up to a backchannel is associated with declining information rate, while speech downstream rebounds to previous rates. Our results provide new insights into long-standing theories of how we respond to fluctuating demands on cognitive resources, and how we negotiate those demands in partnership with others.
\end{abstract}




\end{frontmatter}


Conversation may be the ``mark and cause of friendship''~\cite{aristotle} with a deep evolutionary history~\cite{fire,pauses_universals}, but it is also cognitively demanding. Speakers must translate thought into speech~\cite{slob}, while listeners, in turn, face a now-or-never bottleneck in making sense of what they hear before it vanishes from mind~\cite{christiansen_chater_2016}. The mental workload includes not just language processing~\cite{futrell2015large}, but the theory of mind required to sustain mutual understanding~\cite{dideriksen2019contextualizing,hawkins2019emergence,eyes,hawkins2023partners}.

The challenges of managing this information leave systematic marks on culture and cognition~\cite{special_mt}. If listening is hard, language can evolve so that words that contain more information take longer to say~\cite{steve_length}. Speakers can use pauses and hesitations in a similar fashion~\cite{clark2002using, beattie1979contextual, arnold2003disfluencies}, buying time to call a novel or information-dense word to mind, and then stretching it out when they speak to aid the listener~\cite{aylett_turk}. Both disfluencies (\textit{uh, um}) and backchannels (\textit{mhm, uh-huh}) provide coordinating devices, helping interlocutors communicate more effectively~\cite{fus_coordination,dideriksen2023quantifying,dingemanse2024interjections}. On the grandest scales, the idea that speech rate is at least partially explained by cognitive constraints has prompted the search for a universal (\emph{i.e.}, cross-linguistic) information rate, the product of the interaction between culture and neurocognitive limits, which Ref.~\cite{scirate} most recently estimated at 39 bits/second.

These results depend, for their generality, on a basic result from the theory of computation: the cognitive burden of producing, or processing, a word $w$ ought to be related to its conditional surprise, or Shannon information, defined as $-\log_2{p(w|C)}$: the negative log-probability of $w$ given the context $C$. All other things being equal, $w$ is harder to retrieve or transmit when $p(w|C)$ is small~\cite{uid_people,inf_theory,hale2001probabilistic,levy2008expectation}. Measuring this, however, has proven a challenge because traditional methods for estimating $p(w|C)$ require an exponentially large dataset; Ref.~\cite{steve_length}, for example, was restricted to a context length of just four words. Precisely-defined experiments, or a restriction to particularly clear linguistic structures (\emph{e.g.}, complementizer \emph{that}-mentioning in the foundational work of Ref.~\cite{FLORIANJAEGER201023}) can reduce the space further, at the cost of increasing distance from naturalistic conditions and generalizing to a smaller subspace of use cases.

The emergence of large language models (LLMs) changes this equation. These tools give scientists measures of $p(w|C)$ for context lengths that far exceed what has been possible before, allowing for precise predictions of a word's contextual predictability with long-range context; in this work, for example, our context length is 128 tokens, over half a minute of speech. These LLM measures of $p(w|C)$ correlate closely with reading times, suggesting they reflect human sentence processing to a considerable extent~\cite{shain2022large,de2023scaling,oh2023transformer,wilcox2020predictive}. Here, we combine LLM next-word predictions with a large-scale sample of naturalistic conversation, the CANDOR corpus~\cite{candor}, to examine the real-time effects of fluctuating predictability in everyday talk. Together, these tools enable the simultaneous study of cognitive resource limitations in production, comprehension, and alignment with unprecedented levels of ecological validity.

We report three main results. First, a new estimate of the information rate of spontaneous conversation, less than half of the current ``universal'' bound~\cite{scirate}. Second, we validate a pair of key predictions from the theory of cognitive resource limitations: (1) that high-surprise words are spoken more slowly, both because they are longer and because speakers stretch them out when they occur in high-surprise contexts; (2) that high-surprise words are associated with buying time through both pauses and the lengthening of the preceding word.

Finally, we show how listener backchannels convey information about the information rate. As we show, listeners produce backchannels that respond, on average, to an accumulating sequence of low-surprise events. Backchannels serve as an information-rate feedback mechanism, from listener to speaker; just as the speaker uses word lengths and pauses to regulate the outflow, a listener's backchannel can regulate the inflow.

\section{Results}

\begin{figure}
    \centering
    \includegraphics[width=0.95\textwidth]{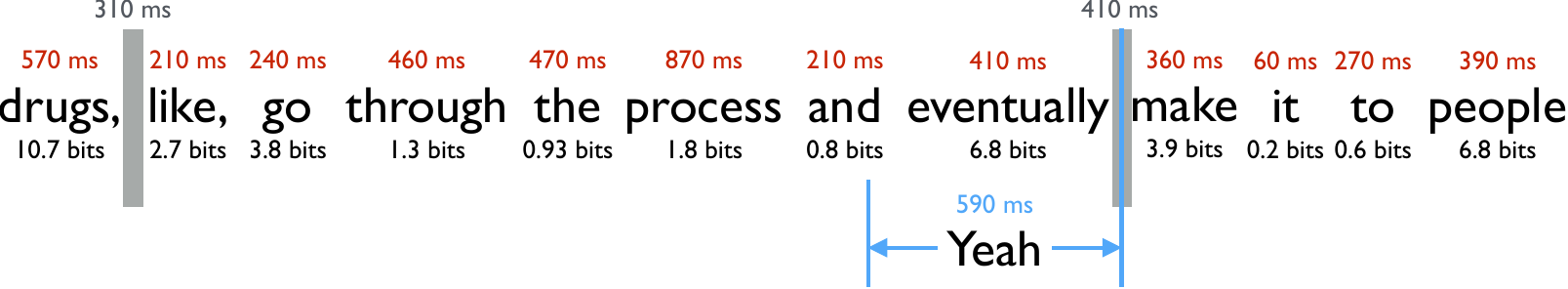}
    \caption{A sample fragment from a processed version of the CANDOR corpus, with LLM-estimated surprise in bits and AWS-estimated word lengths in milliseconds. Grey bars mark pauses greater than 10 ms. For example, the word ``like'' follows the word ``drugs'' after a pause of 310 ms; the word lasts 210 ms, and has surprise $2.7$ bits. In this example, the second speaker provides a backchannel (``yeah'') that begins while the first speaker is finishing the word ``and'', and lasts 590 ms, through into the pause following ``eventually''. The LLM context window is 128 tokens.}
    \label{sentence}
\end{figure}

We present our results in four sections; see~\ref{methods} for details on our materials and methods. Fig.~\ref{sentence} provides a simple example of what our data looks like.

\subsection{The Information Density of Conversation}

\begin{figure}
    \centering
    \includegraphics[width=0.5\textwidth]{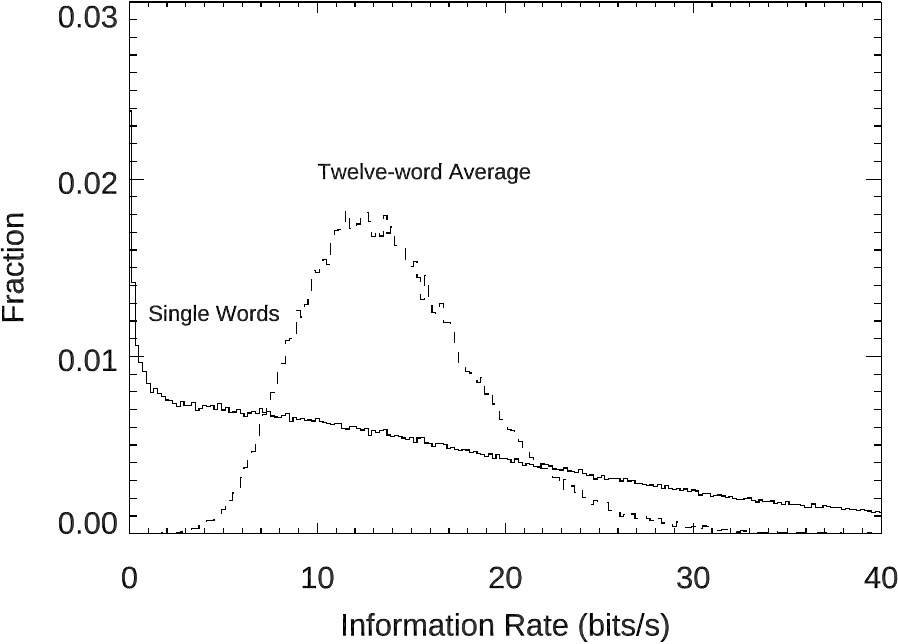}
    \caption{The distribution of information rates in continuous conversational English, CANDOR corpus (659,496 words, 54,958 sequences). Solid line: information density of single words; dashed line: average density over the course of twelve consecutive words. \label{dist}}
\end{figure}

Our measures of information from the Open LLama LLM, combined with timing data from AWS, give us a new estimate of the information rate of conversational English. In uninterrupted sequences of twelve words---a length long enough to include basic syntactical structures such as noun and verb phrases, even when accompanied by disfluencies---this is $4.04\pm0.01$ bits/word, produced at $13.21\pm0.04$ bits/second ($N=54,958$ sequences). As shown in Fig.~\ref{dist}, the distribution at the word-by-word level is broad---there are a significant number of highly predictable moments in any conversation, for example, and more than ten percent of words carry less than half a bit of information. Much speech is filler.

Our 13 bits/second result is significantly lower than the ``universal'' rate for human languages, of 39 bits/second, found by Ref.~\cite{scirate}. The reason is that we can account for context: a listener who attends to the speaker will often (though not always) have a lower surprise than one who hears the same word out of context. The word ``protein'' is relatively rare, and thus has high baseline surprise, but when it appears in the sentence ``people need to eat more fish, which provides healthy protein'', the surprise is much lower. 

The increased ecological validity that comes from capturing the effects of context has costs; among other things, our measurements depend on the nature of the underlying LLMs---none of which were trained on purely conversational text. If humans are better at predicting in context than the LLM, their surprise should be lower and (by Jensen's Inequality) our estimate becomes an upper bound relative to the better-tuned model that would be possessed by a human speaker. Human speakers may also use cues the model does not have access to---gestures, visual context, and other non-verbal cues---to better predict what someone will say; again, this preserves our estimate as an upper bound.

Conversely, the integrity of our estimate as an upper bound is threatened if LLMs are better at predicting words in conversational context than humans are---as might happen if the listener's attention flags, or if they use their additional context in a biased fashion that makes their model worse.

We emphasize that our 13 bits/second result concerns the average information rate in a few seconds of unstructured conversation, and not a limit on the information rate of an arbitrary stretch of human speech. Even within our sample, as can be seen in Fig.~\ref{dist}, there is a great deal of variability; consistent with Ref.~\cite{scirate}, the variance in information rates ($\sigma\approx4.9$ bits/s) is significant, and some seconds of speech do much more than others.

\subsection{Presenting Surprise: Efficient Coding}
\label{presentation_section}

A major prediction of resource constraint models is that words with higher surprise should take longer to say. Two processes contribute to this relationship. First, words that are produced more often are shorter~\cite{zipf2013psycho,sigurd2004word}. Under a naive Shannon model, for example, an efficient code would lead the length of a word to increase linearly with surprise~\cite{steve_length}. Second, we expect speakers to slow down when presenting a word that is more surprising for their listener given context, an effect related to the smooth signal redundancy hypothesis~\cite{aylett_turk}. Work using next- and previous-word predictability finds that less predictable words are pronounced more slowly~\cite{bell2003effects, bell2009predictability,pluymaekers2005articulatory}; our work revisits these results using the far wider contexts enabled by the LLM.

\begin{figure}
    \centering
    \includegraphics[width=0.5\textwidth]{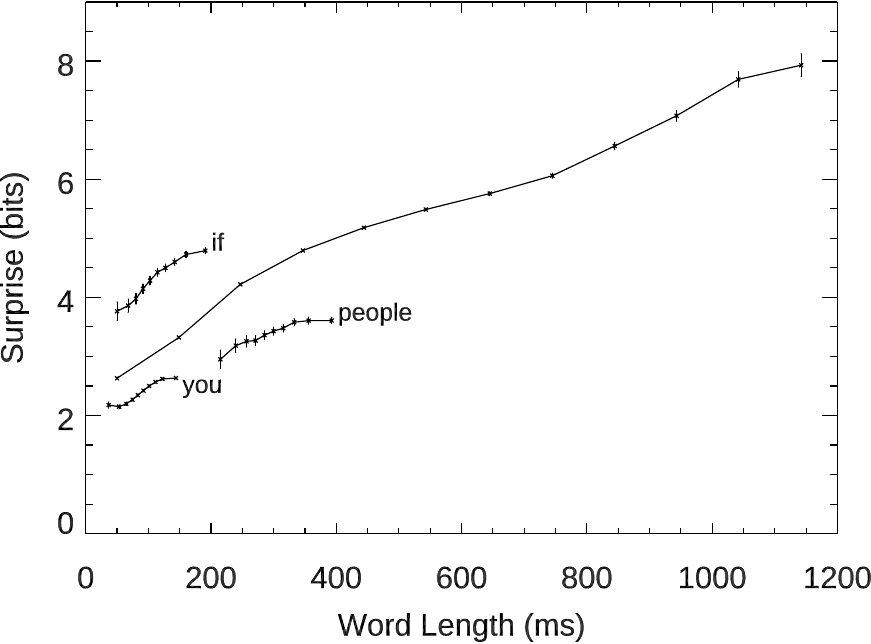}
    \caption{Words with higher surprise take longer to say. This holds not only in the aggregate (solid line), but even for the same word: when a word appears in a higher-surprise context, it tends to be said more slowly than it is in a low-surprise context. Dashed lines show three examples of this effect for common words: ``you'' (N=14,150), ``if'' (N=2701), and ``people'' (N=2534), binned by decile in timing.}
    \label{rate}
\end{figure}
As shown in Figure~\ref{rate}, this is indeed what we find. The most rapidly produced words are, on average, lower surprise; a simple linear model with speaker- and partner-level random effects finds a coefficient $5.50\pm0.02$ bits/s ($r^2$=0.10, N=567,762, $t$=+249); a word that is 100 ms longer than the conversation average has 0.55 bits of additional surprise.

A prediction of the second process is that we should continue to see the length--surprise relationship when the same word appears in more or less predictable contexts. This is also clearly detectable: after including word-level random effects (and speaker and partner effects as before), we find $0.97\pm0.03$ bits/s ($t$=+37, $r^2$=0.002). Excluding the 100 most frequent words increases the effect ($1.42\pm0.05$ bits/s; $t$=+27; N=191,236, $r^2$=0.004). 

Our effect sizes are large, but our $r^2$ for these relationships is smaller than what psychologists are used to from experimental studies of other forms of behavior. They are, however, similar in magnitude to prior synoptic observational studies of language processing (\emph{e.g.}, Ref.~\cite{steve_length}), and with the residuals explained by information theoretic effects in more controlled studies of conversation (\emph{e.g.}, Ref.~\cite{FLORIANJAEGER201023,aylett_turk}). There are many things that affect presentation and production, including both generic effects (\emph{e.g.}, some common, low-surprise English words, such ``people'', can still take a long time to say), semantic ones (\emph{e.g.}, the prolongation of a word for emphasis), situational ones (\emph{e.g.}, the dynamically-generated rapport between speakers can cause shifts in speech rate over the course of the conversation), and aspects of the context not captured in $C$ (\emph{e.g.}, visual information, implicit shared context, or references to much earlier information). There are also observational limitations; most notably for us, the variance in timing errors which, at the word level, is comparable with the overall variance. We return to this question in the discussion.

\subsection{Producing Surprise: Computational Resource Constraints}
\label{production_section}

Speakers face two challenges: they must bring to mind the word they want to say (information ``retrieval''), and they must say it (information ``presentation''). The previous section focused on the constraints of presentation, which come from both the structure of the English language, and (as our word-level model shows) the dynamical adjustments that speakers can make on the fly.

Retrieval is a separate matter, and equally subject to computational resource constraints. A speaker can not begin to speak until they have some idea of the word they wish to say. Figuring out what to say, in turn, should be harder when the word to be retrieved has high surprise; a Shannon-efficient retrieval, for example, that dynamically adapts to context will incur a time cost proportional to surprise.

A prediction of the resource constraint model is thus that speakers need to buy time for high-surprise words before they begin saying them~\cite{goldman1958predictability,beattie1979contextual,harmon2015studying}. There are several natural mechanisms for this. Speakers might draw out the preceding word (elongation), pause before beginning the word (hiatus), or use a disfluency such as ``uh'', ``um'', or ``like''.

These three disfluencies alone are exceedingly common, and account for 6\% of speech in our corpus; while ``like'' has non-disfluent uses (\emph{e.g.}, as a verb, a comparative, or to introduce speech), approximately 78\% of its occurrences in our corpus are filler (1426 sampled ``likes'' in 100 conversations).

A regression model with speaker and partner fixed effects shows all three effects in play. Elongation and hiatus are relatively weak effects, predicting $0.36\pm0.02$ bits/s ($t=+15$) and $1.06\pm0.04$ bits/s ($t=+29$), respectively; \emph{e.g.}, an additional 150 ms pause predicts an additional $0.16$ bits of surprise. Disfluencies are more powerful; a disfluency just prior predicts an additional $1.71\pm0.02$ bits of surprise. This holds equally, if not more strongly, after accounting for the effect the disfluency has on the LLM's ability to predict future speech (see Materials and Methods). As in the previous section, our effect sizes are large---a disfluency, when it appears, can explain more than a quarter of the future surprise variance on average---but our (marginal) $r^2$ is low (0.02 for retrieval alone; 0.11 for for presentation and retrieval together). We return to this question in the discussion.

Interestingly, we do not see ``exhaustion'' effects: producing a high surprise word in fact (very slightly) increases the likelihood of producing another high surprise word immediately after ($0.016\pm0.001$ bits/bit, $t=+12$), capturing, if anything, higher-level effects of the ebb-and-flow of interesting material, and controlling for surprise does not significantly alter the results above.

\subsection{Synchronizing Minds: Backchannels and Communication Constraints}
\label{bc_section}

Retrieving and presenting words is necessary for a successful conversation, but it is far from sufficient. For communication to work, speaker and listener must share reasonably similar mental models of the world, and update them dynamically in response to each other. It is not enough for me to speak: I also must find out if you are making sense of what I say, and whether you can handle more. This kind of interactivity can do more than just reassure; it also enables the speaker to leverage the listener's signals to achieve higher rates of transmission~\cite{bc_theory}.

One mechanism for this is the use of backchannel communication, where the listener communicates to the speaker without interrupting the speaker's flow. These include single-word interjections shown in Figure~\ref{sentence}; the ``uh-huh''s, ``yeahs'', and ``mhmm''s that provide a more-or-less encouraging backdrop to the speaker's speech. Explicit verbal backchannels appear in the CANDOR corpus in abundance; they occur an average of every 23 seconds, and 20\% of the turns in a conversation consist of a listener who backchannels a single word before returning control of the conversation to their partner.

Our hypothesis is that backchannels help with two key synchronizing functions: (1) they alert the speaker to the fact that the listener finds the current sequence of words predictable; (2) they encourage the onset, or resumption, of novel material. Function (1) suggests that we should see decreasing surprise in the lead-up to a backchannel, while Function (2) suggests that we should see increased surprise following it.

\begin{figure}
    \centering
    \begin{tabular}{c}
    \includegraphics[width=0.5\textwidth]{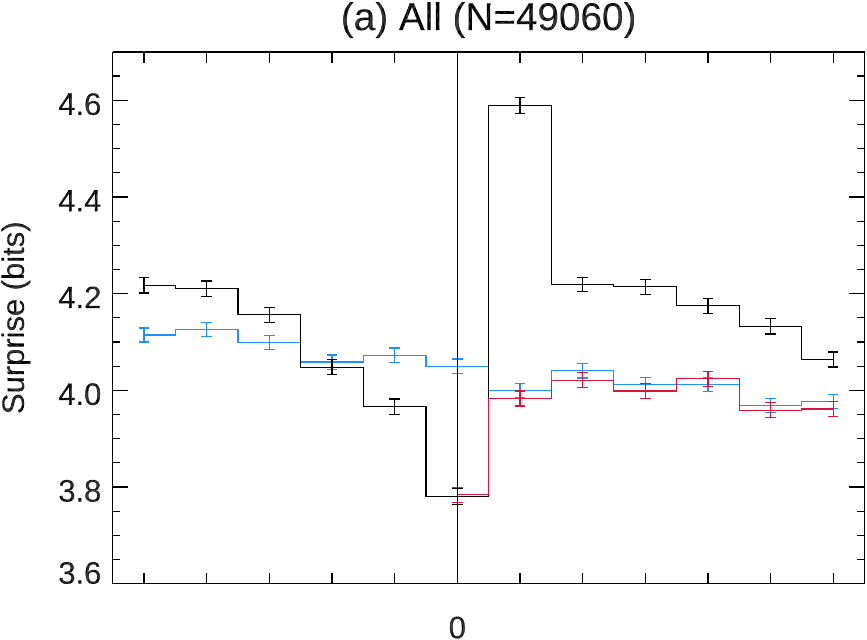} \\
    \includegraphics[width=0.5\textwidth]{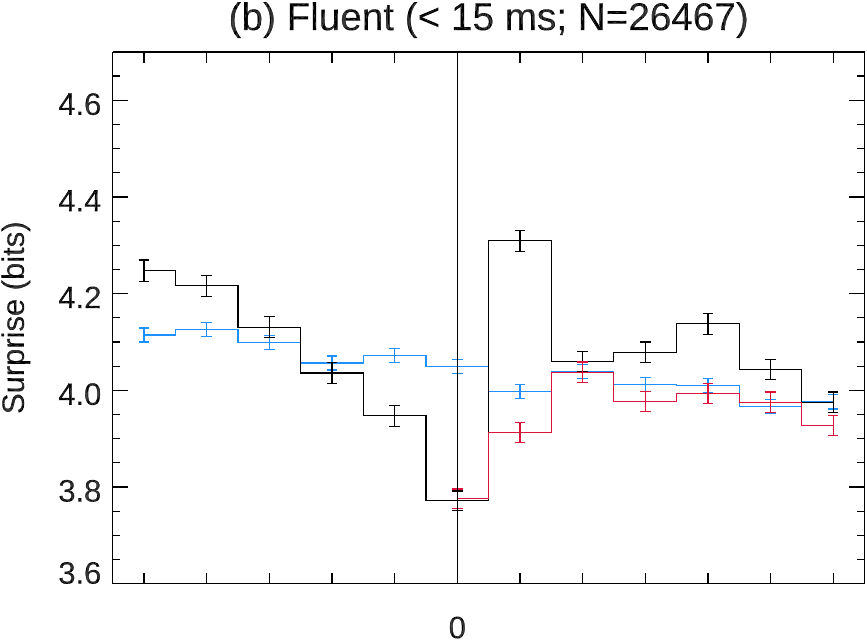}
    \end{tabular}
    \caption{The timeseries of surprise for continuous back-channeled speech (solid black line) relative to the base model (blue line), and relative to a sample from the base model that matches the surprise distribution at backchannel (red line), with standard errors. The backchannel occurs at word position zero (\emph{e.g.}, in Fig.~\ref{sentence}, the word ``and''). All: all backchannels; Fluent: a backchannel that follows a word with a gap less than 15 ms.}
    \label{bc}
\end{figure}

A simple way to look for these effects is to compare the surprise time-series for backchanneled speech to the baseline model. This is shown in Figure~\ref{bc}. Position zero in this time-series corresponds to the word that overlaps with the backchannel: \emph{i.e.}, the backchannel either occurs while the word is being spoken, or in the gap between that word and the next, if a gap exists. The blue line corresponds to the baseline---\emph{i.e.}, sequences that do not contain a backchannel.

Two effects are immediately visible in Fig.~\ref{bc}(a). First, the speaker's surprise drops continuously (on average) in the words preceding a backchannel (black line), an effect not seen in ordinary, unbackchanneled speech (blue line). Second, the surprise rebounds, to above baseline levels, after. Fig.~\ref{bc}(b) shows the case of fluent speech, where the word overlapping the backchannel follows immediately, without a gap, where the same effect is seen.

It is natural to ask if the rise in surprise after the backchannel is driven simply by the fact that (1) backchannels tend to occur at moments of low relative surprise, and (2) moments of low relative surprise might tend to be followed a rebound to higher levels. One way to check for this effect is with a subsample from the base model that matches the surprise distribution just prior to the back-channel; comparing the red and black lines shows that the rebound effect is not just an artifact of an anomalously low-surprise word position.

In interpreting Fig.~\ref{bc}, the relative timing of the speaker's words and the listener's backchannel matters a great deal; particularly in the case of fluent speech, where one word flows seamlessly into the next without a clear silence, this can be hard to achieve. We hand-checked twenty randomly-selected back-channels from our data, comparing the AWS-provided timestamps against the raw waveform using the Audacity editing software. We found that the precise timing is somewhat biased; in particular, in about seven out of the twenty cases checked by hand, the backchannel actually falls on the next word, \emph{i.e.}, in these cases, the zero point of Fig.~\ref{bc} should be shifted one unit to the right; in the other 13 cases, the backchannel is correctly located.

A rebound that occurs either with, or one word after, the backchannel, imposes a tight timing constraint: even a fast-witted speaker will have difficulty dynamically altering the structure of their sentence in fractions of a second. The sample fragment in Fig.~\ref{sentence} happens to provide an example of this effect---the ``yeah'' begins just as the low-surprise ``and'' (0.8 bits) is finishing, and (just) before ``eventually'' (6.8 bits) begins. Fig.~\ref{bc}(b) suggests that this is a repeatable phenomenon.

These results suggest at least partial use of prediction~\cite{corps2022overrated}: a listener not only responds to the decrease in surprise, but also correctly anticipates the point at which the speaker will start to generate more novel material. Such a prediction is possible; our method can also measure the predicted information (\emph{i.e.}, the expectation value of $-\log_2 p(w|C)$ over all words $w$, not just the actual one), and we see the same pattern of rise in the predicted surprise after the zero-point for fluent speech.

Our results are consistent with models of language production~\cite{pickering2013integrated,levinson2015timing,levinson2016turn,Corps2018230} that place a premium on both forward prediction and planning---in this case, not only the listener anticipating the correct moment to backchannel, but also the speaker planning their speech under the assumption that the backchannel will, indeed, occur where it ought. Such constraints would not operate when pauses are longer; in these cases---a little less than half our sample---dynamical adaptation by the speaker is more likely. 


\section{Discussion}

The cognitive demands of conversation serve as an excellent proving-ground for generalizable theories of how we think. To this end, the CANDOR corpus allows us to validate basic hypotheses about how cognitive load, as measured by Shannon information, shapes the production and processing of language. It also provides new insight into how we solve the problem of interactivity. Backchannels can provide the speaker with meta-information---facts not about the world, but about the listener's epistemic surprise---and are associated, in turn, with shifts in the speaker's downstream behavior.

One general finding, which accords with a range of prior results, is that while cognitive resource limitations are certainly detectable with Shannon information on average, there is a great deal of variability, particularly on short timescales, that is unaccounted for by purely Markovian information-theoretic demands. Even taking into account measurement errors in timing, much about how people time their utterances remains unexplained.

The accidents of language evolution and the presence of unaccounted-for context go part of the way towards explaining this shortfall. A deeper reason, however, is that language processing is multiscale~\cite{christiansen_chater_2016}, which means that Shannon information only captures one dimension of computational resource costs. Consider, for example, the work of Ref.~\cite{futrell2015large}, which studied the cognitive load of dependency trees; the sentence ``John threw out the trash'' is more easily processed than ``John threw the trash out'', because ``trash'' and ``out'', in the latter case, are further from the relevant verb. If the two sentences are equally probable given context, however, they will have equal Shannon information. Surprise considers the costs of retrieval and transmission, but not the computational costs of transforming for either the speaker or the listener.

The costs of processing go beyond syntax. Given the context ``you did so badly on this exam that your grade is'', a very probable next word may well be the failing grade ``F'', easily retrievable from memory. Many teachers, however, will incur excess cognitive load in speaking at this stage, double-checking their thoughts because of the high stakes of a false positive. Communication is more than information transmission, and we can hesitate even before the nearly inevitable, both good and bad~\cite{templeton2023long,fraundorf2008dimensions}.

Taken together, our results are double-sided. The large effect sizes for each of our studies show that computational load matters a great deal, and they reveal new ways human communication is cooperatively managed by resource-limited minds. At the same time, they suggest that simple Shannon measures fall short of capturing the full range of cognitive activity that brains undertake. This suggests a new role for theorists at the intersection of linguistics and the theory of computation, who can provide generalizable frameworks for studying the true computational complexity of this human universal.

\section{Acknowledgements}

We thank Gus Cooney, Gülce Kardeş, and Robert Hawkins for helpful comments and advice on this work.


\section{Methods}
\label{methods}

Our base sample is a subset of the full CANDOR corpus: audio from 1469 English-language Zoom conversations~\cite{candor} retranscribed by the CANDOR team using the AWS (January 2024) model in multichannel model, keeping filler words. For each conversation, we sample windows of twelve words at random from the conversation (excluding the first three and last three turns); for each word $w$ in the window, we compute both the surprise conditional on the context, $-\log_2 p(w|C)$, and the expected surprise (conditional entropy) over the full vocabulary $W$, $-\sum_{i\in W} p(i|C) \log_2 p(i|C)$. 

The prior context $C$ consists of the 128 tokens of the interlocutor's speech preceding the word $w$ in question. We ran the Open Llama 7B model\footnote{\url{https://github.com/openlm-research/open\_llama}}, using llama.cpp with 4-bit integer quantization\footnote{\url{https://github.com/ggerganov/llama.cpp}} so that it can run on consumer hardware. We experimented with other levels of quantization, with increasing the window size to 256 tokens, and with model weights from Llama 2 and Mistral 7B, finding only minor differences. We do not include surprise for the small number of numerals (``34'', etc) and special characters (such as ``\$'') that appear in the transcript.

Our base sample consists of 115,531 windows and 1,386,372 words; from that base sample, we construct a ``continuous'' subsample where the window consists of uninterrupted speech (\emph{i.e.}, the twelve words are from the same speaker). The continuous subsample has 54,958 windows (659,496 words); this is a more conservative sample for studying information constraints, because turn changes within the window can introduce anomalous levels of surprise if the other speaker induces a topic change. Extending this window further does not strongly affect our results. In our regression analyses of Sections~\ref{presentation_section} and~\ref{production_section}, we drop words where the length of the word, or the prior word, including pauses, is longer than two seconds, although this does not strongly affect our results.

In addition to our base sample, we also construct a separate sample of windows that include a backchannel; a backchannel is defined as a speaker turn that includes only one word: ``yeah" (by far the most common, 48\% of the sample), ``mhmm" (15\%), ``okay" (8\%), ``oh" (8\%), ``right" (7\%), ``uh" (3\%), ``yep" (3\%), ``yes" (3\%), ``wow" (2\%), or ``uhhuh" (2\%)---the vast majority are what Ref.~\cite{schegloff1982discourse} calls continuers~\cite{bangerter2003navigating}; the overall backchannel rate is consistent with those found in other studies~\cite{knudsen2020forgotten}. The window is centered on the backchannel, and same calculations are done as for the base. Our backchannel sample consists of 117,998 windows and 1,415,976 words; we also construct a continuous subsample, where at least six words both before and after the backchannel are from the same turn; this contains 49,060 windows and 588,720 words. Figure~\ref{sentence} provides an example of how this works, along with the relevant variables. 

In our study of retrieval, we looked at the use of the disfluencies ``um'', ``uh'', and ``like''. It is possible that the introduction of these words makes the LLM a worse predictor of the next word, inducing an artificial association. We checked for this by re-running our LLM model on a subset of data ($N=558$) that included disfluencies; the change in LLM surprise on the next word from \emph{removing} the prior disfluency is, on average, positive, $1.7\pm0.2$; this indicates that our finding is, if anything, an underestimate of the full effect of how a disfluency can ``buy time''.

\clearpage
 \bibliographystyle{elsarticle-num-names} 
 \bibliography{main}





\end{document}